\documentclass{article}

\usepackage{PRIMEarxiv}

\usepackage[utf8]{inputenc} 
\usepackage[T1]{fontenc}    
\usepackage{hyperref}       
\usepackage{url}            
\usepackage{booktabs}       
\usepackage{amsfonts}       
\usepackage{nicefrac}       
\usepackage{microtype}      
\usepackage{lipsum}
\usepackage{fancyhdr}       
\usepackage{graphicx}       
\graphicspath{{media/}}     
\usepackage{algorithm}
\usepackage{algorithmic}
\usepackage{multirow}
\usepackage{array}
\usepackage{amsmath}

\title{Flame-state monitoring based on very low number of visible or infrared images via few-shot learning
}

\author{
   Ruiyuan Kang \\
  Department of Mechanical Engineering \\
  Khalifa University \\
  Abu Dhabi, UAE\\
  \texttt{ruiyuan.kang@ku.ac.ae} \\
       \And
  Panos Liatsis \\
  Department of Electrical Engineering and Computer Science \\
  Khalifa University \\
   Abu Dhabi, UAE\\
  \texttt{panos.liatsis@ku.ac.ae} \\
   \And
  Dimitrios C. Kyritsis \\
  Department of Mechanical Engineering \\
  Research and Innovation Center on CO$_2$ and Hydrogen \\
  Khalifa University \\
   Abu Dhabi, UAE\\
  \texttt{dimitrios.kyritsis@ku.ac.ae} \\
}

\begin{document}
\maketitle

\begin{abstract}
 The current success of machine learning on image-based combustion monitoring is based on massive data, which is costly even impossible for industrial applications. To address this conflict, we introduce few-shot learning in order to achieve combustion monitoring and classification for the first time. Two algorithms, Siamese Network coupled with k Nearest Neighbors (SN-kNN) and Prototypical Network (PN), were tested.  Rather than utilizing solely visible images as discussed in previous studies, we also used Infrared (IR) images. We analyzed the training process, test performance and inference speed of two algorithms on both image formats, and also used t-SNE to visualize learned features. The results demonstrated that both SN-kNN and PN were capable to distinguish flame states from learning with merely 20 images per flame state. The worst performance, which was realized by PN on IR images, still possessed precision, accuracy, recall, and F1-score above 0.95. We showed that visible images demonstrated more substantial differences between classes and presented more consistent patterns inside the class, which made the training speed and model performance better compared to IR images. In contrast, the relatively low quality of IR images made it difficult for PN to extract distinguishable prototypes, which caused relatively weak performance.  With the entrire training set supporting classification, SN-kNN performed well with IR images. On the other hand, benefitting from the architecture design, PN has a much faster speed in training and inference than SN-kNN. The presented work analyzed the characteristics of both algorithms and image formats for the first time, thus providing guidance for their future utilization in combustion monitoring tasks.
\end{abstract}

\keywords{Combustion state monitoring\and Few-shot learning\and visible images\and infrared images}

\section{Introduction}\label{1introduction}
Combustion is a process that appears in the majority of current energy conversion technologies and entails a combination of complicated chemical kinetics with hydrodynamically unstable flows that evolves in occasionally complicated geometries.  In order to assure the safety, quality, controllability, and recently popular, environmental friendliness of combustion, combustion monitoring is an indispensable engineering tool, especially given the fact that accurate, time-resolved, direct numerical simulation of the phenomenon is way above current computational capabilities and will remain so for the foreseeable future.

Although laser-based diagnostics ~\cite{kohse-hoinghausAppliedCombustionDiagnostics2002,eckbreth1996laser,kangIntelligenceComplexityMachine2022} have achieved tremendous success in the field of combustion monitoring, passive optical gas imaging ~\cite{hernandezFlameImagingDiagnostic2008} is gradually attracting attention as a potential combustion monitoring tool. Using just a portable camera rather than a laser-based system is, of course, more convenient, and less costly. Nevertheless, the information delivered by optical images is implicit, non-quantitative and potentially elusive, although experienced on-site engineers may have developed the situation-specific sense to infer (approximate) information on flame structure and possibly emissions from images.

Recently, the advance of AI has been progressively alleviated the dependence on the intuition of skilled and experienced individuals and empowered multiple combustion monitoring tasks, e.g., monitoring combustion instability ~\cite{baiMultimodeMonitoringOxyGas2017}, classifying fuel composition ~\cite{zhouSupportVectorMachine2014,hanCombustionStabilityMonitoring2020},  as well as propensity towards soot formation ~\cite{pinoSootPropensityImage2018}. In general,   the methodologies utilized in these studies can be divided into three categories (Table ~\ref{table1}): (1) Coupling feature engineering with classical supervised learning; (2) Coupling representation learning with classical supervised learning; (3) Deep supervised learning.

\begin{table}[]
\caption{Brief categories of work on AI-assisted image-based combustion state monitoring}
\label{table1}
\resizebox{\textwidth}{!}{%
\begin{tabular}{|l|l|}
\hline
\multicolumn{1}{|c|}{\textbf{Methodology}} & \multicolumn{1}{c|}{\textbf{References}} \\ \hline
Feature engineering + classical supervised learning & \begin{tabular}[c]{@{}l@{}}Zhou, et al.~\cite{zhouSupportVectorMachine2014},Bai, et al.~\cite{baiMultimodeMonitoringOxyGas2017},Pino, et al.~\cite{pinoSootPropensityImage2018},Wang, et al.~\cite{wangPatternRecognitionMeasuring2020}, Chen, et al. ~\cite{chenBurningConditionRecognition2020}, \\ De Santana, et   al.~\cite{desantanachuiFuzzyInferenceOil2019}, Li. et al. ~\cite{liFlameImagebasedBurning2012}, Hauser. et al. ~\cite{hauserRealtimeCombustionState2016}, \\ Chui, et al. ~\cite{chuiDynamicsFlameImages2020}, Chen,   et al. ~\cite{chenMonitoringCombustionSystems2010,chenGaussianProcessRegression2013}, Sun. et al. ~\cite{sunConditionMonitoringCombustion2013}\end{tabular} \\ \hline
Representation learning + classical supervised learning & Wang. et al. ~\cite{wangPatternRecognitionMeasuring2020}, Han, et al.~\cite{hanCombustionStabilityMonitoring2020,hossainCombustionConditionMonitoring2019}, Akintayo, et al. ~\cite{akintayoEarlyDetectionCombustion2016}, Qiu, et al.   ~\cite{qiuUnsupervisedClassificationMethod2019} \\ \hline
Deep supervised learning & Wang, et al.~\cite{wangDeepLearningBased2017}, Zhu. et al. ~\cite{zhuConvolutionalNeuralNetwork2019}, Li, et al.~\cite{liPredictingCombustionState2019}, Choi, et al.~\cite{choiCombustionInstabilityMonitoring2020} \\ \hline
\end{tabular}%
}
\end{table}

In the first and second categories the idea is to extract low-dimensional features in order to represent high dimensional images, thus avoiding the problem named as dimentionality curse ~\cite{trunkProblemDimensionalitySimple1979}. However, in specific implementations, these two categories of methods have intrinsic differences.

In particular, for the first category, the features are extracted by multiple processes explored and designed by researchers themselves through their domain knowledge. For example, Refs.~\cite{chenMonitoringCombustionSystems2010,chenGaussianProcessRegression2013} utilized Principal Component Analysis (PCA) ~\cite{abdiPrincipalComponentAnalysis2010} to compress the images into low dimensional vectors, i.e., features.  Based on that, Ref. ~\cite{baiMultimodeMonitoringOxyGas2017,sunConditionMonitoringCombustion2013} used upgraded kernel PCA ~\cite{scholkopf1997kernel}, which can further extract nonlinear features to describe images. Besides, Ref.~\cite{wangPatternRecognitionMeasuring2020,chenBurningConditionRecognition2020} used statistical parameters, such as norms and kurtosis, to describe images. In addition, operations of image processing were often utilized, such as Gabor filter  ~\cite{liFlameImagebasedBurning2012}, histogram of oriented gradients ~\cite{hauserRealtimeCombustionState2016}, and Ibrahim time-domain transform ~\cite{chuiDynamicsFlameImages2020}. Such manual feature engineering is time-consuming, since massive combinations need to be explored in order to find a proper way to describe images.

To tackle this shortcoming, in the second category the features are extracted automatically through machine learning algorithms. For example, Wang et al. ~\cite{wangPatternRecognitionMeasuring2020} utilized self-organization map ~\cite{kohonenSelforganizingMap1990}, a competitive learning algorithm, to learn low dimensional features, Ref.~\cite{hanCombustionStabilityMonitoring2020,akintayoEarlyDetectionCombustion2016,qiuUnsupervisedClassificationMethod2019} utilized diverse versions of auto encoder to embed images ~\cite{dong2018review} into compact features; Hossain, et al. ~\cite{hossainCombustionConditionMonitoring2019} utilized Generative Adversarial Network(GAN) ~\cite{creswellGenerativeAdversarialNetworks2018} to learn more robust features in an adversarial way. Once informational features are extracted, both categories have no difference in utilizing classical supervised learning algorithms to realize combustion monitoring, i.e., mapping features to diverse classes, such as combustion states. The classical supervised learning algorithms have been attempted include Support Vector Machine (SVM) ~\cite{nobleWhatSupportVector2006}, Multi-Layer Perceptron( MLP) ~\cite{gardnerArtificialNeuralNetworks1998}, Gaussian process ~\cite{williams2006gaussian}, etc.

The third category encapsulates the feature engineering and classification into an integral process, that is termed as end-to-end learning. In such a process, the features learned are classification oriented, which is more targeted. In general, this idea is realized by convolution neural networks~\cite{al-saffarReviewDeepConvolution2017}, while some works also coupled recurrent neural network ~\cite{liPredictingCombustionState2019,choiCombustionInstabilityMonitoring2020,tsoiRecurrentNeuralNetwork1998} in order to utilize the temporal information from image sequences.

These works have contributed significantly to the development of image-based combustion monitoring. But, the fact of the matter is that a tremendous amount of data, at least thousands of images, are needed for training a machine learning model, especially, a deep learning one. This causes (1) massive cost in terms of time and computational resources for the purpose of collecting, organizing, and labelling images; (2) very costly training of a convergent model in terms of both time and resources.  These two requirements conflict with industrial practice because industrial applications of combustion monitoring cannot often provide sufficient and good-quality data, since abnormal operation states that combustion monitoring mainly pays attention to are often transient, low-probability, and deliberately avoided, thus, very few available data can be collected. and moreover, industrial practice very often cannot afford the long time needed in order to train networks based on data-rich images. Therefore, to alleviate the conflicts, an intelligent system which can learn efficiently from very limited data emerges as a pressing industrial need.

In addition to the consideration of algorithms, to the authors’ best knowledge, previously published studies have only utilized images in the visible band (Table ~\ref{table1}). However, IR images may be also be a potentially instrumental alternative, since many fuels of interest, as well as products of combustion have strong IR absorption/emission bands (hydrocarbons, ammonia, CO$_2$, CO, and NO$_x$) ~\cite{kangEmissionQuantificationPassive2022,vollmerInfraredThermalImaging2017}. Meanwhile, the major atmospheric gases (nitrogen and oxygen) do not have IR activity and can therefore not disturb the signal. Therefore, it is also theoretically appealing to utilize IR images for combustion monitoring. 

In this paper, we introduce two few-shot-learning-based combustion monitoring algorithms, namely, Siamese Network with k nearest Neighbors and Prototypical Network, which we utilize in order to realize combustion-state classification. Rather than solely training and testing these algorithms on visible images, we also assess the feasibility of utilizing IR images. In doing so, we aspire to contribute the following:
\begin{enumerate}
    \item Introduce few-shot learning into the field of combustion monitoring for the first time, which can significantly save the data needed for training a deep learning model, and thus massively reduce the time needed for training.
    \item Assess and prove the feasibility of utilizing IR images to realize combustion monitoring task for the first time.
    \item Compare the performance and analyze the characteristics of both algorithms on both image formats (visible and IR), which provides the guidance for further utilization of the examined combinations of algorithms and image formats.
\end{enumerate}

\section{Methodology}
\subsection{Experiment setup and data acquisition}
\begin{figure}[hbt!]
\centering
\includegraphics[width=.8\textwidth]{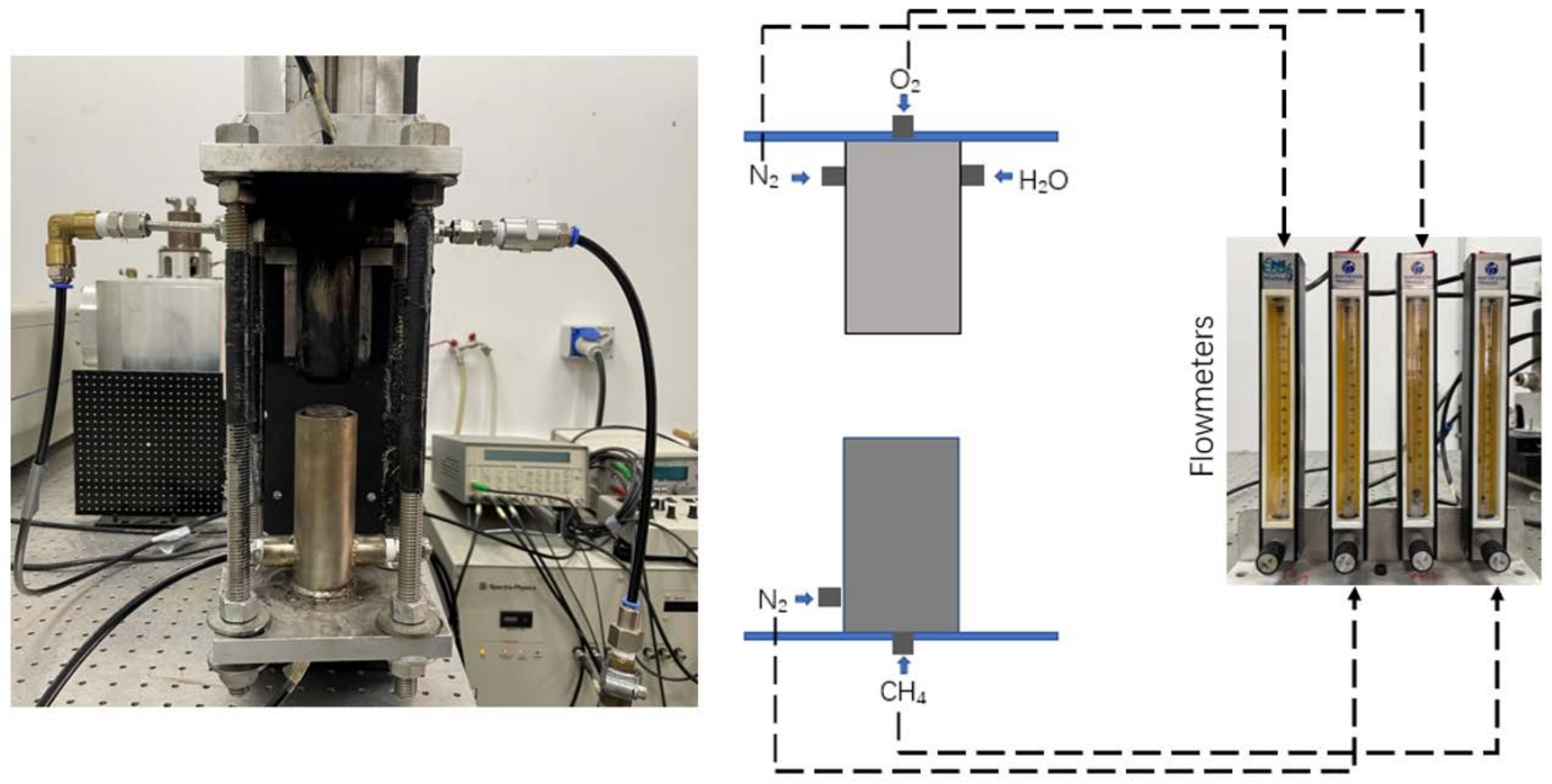}
\caption{Counterflow burner and schematic of the experimental set up}
\label{fig1}
\end{figure}

IR and visible images of flames were acquired during combustion experiments in the counterflow burner built in Khalifa University ~\cite{farraj2017laminar}. Figure ~\ref{fig1}  shows the configuration of the burner, O$_2$ and CH$_4$ were metered to the upper and lower nozzle respectively.  H$_2$O was supplied to the upper nozzle for cooling. In addition, N$_2$ is metered through both nozzles in order to provide the capability to stabilize the flame through controlling the imposed strain without affecting the heat release. We used two cameras to observe the flames. (1) FASTEC TS-5 high-speed CMOS camera, with sensitivity in the is 350 - 950 nm range (which includes the visible band) and a download rate of 100 frams per second (FPS), and (2) FLIR E95 IR camera with sensitivity in the mid-IR, specifically 7 - 15 $\mu$m and a download rate of 30 FPS.

Six different compositions of the counterflowing streams were set, the flames consequently generated were regarded as six combustion states, which are also termed as “classes” in the context of machine learning. The flow rates of nitrogen, oxygen, and methane were measured with rotameters with an estimated error of 5\% according to the specifications of the manufacturer. The mass-flow rates of each gas for each of the flame classes are shown in Table ~\ref{table2}.

\begin{table}[]
\centering
\caption{Mass flow rates of gases in the stream of reactants in g/s}
\label{table2}
\begin{tabular}{|c|c|c|c|c|}
\hline
\textbf{Flame class} & \textbf{N$_2$ in O$_2$} & \textbf{N$_2$ in CH$_4$} & \textbf{O$_2$} & \textbf{CH$_4$} \\ \hline
Class 1 & 1.03 & 0.52 & 1.82 & 0.77 \\ \hline
Class 2 & 1.14 & 0 & 2.01 & 1.19 \\ \hline
Class 3 & 2.72 & 0 & 2.68 & 0.92 \\ \hline
Class 4 & 0 & 0 & 0.95 & 0.52 \\ \hline
Class 5 & 2.02 & 1.19 & 1.25 & 0.78 \\ \hline
Class 6 & 1.11 & 1.04 & 1.28 & 1.25 \\ \hline
\end{tabular}
\end{table}

\begin{figure}[hbt!]
\centering
\includegraphics[width=1.\textwidth]{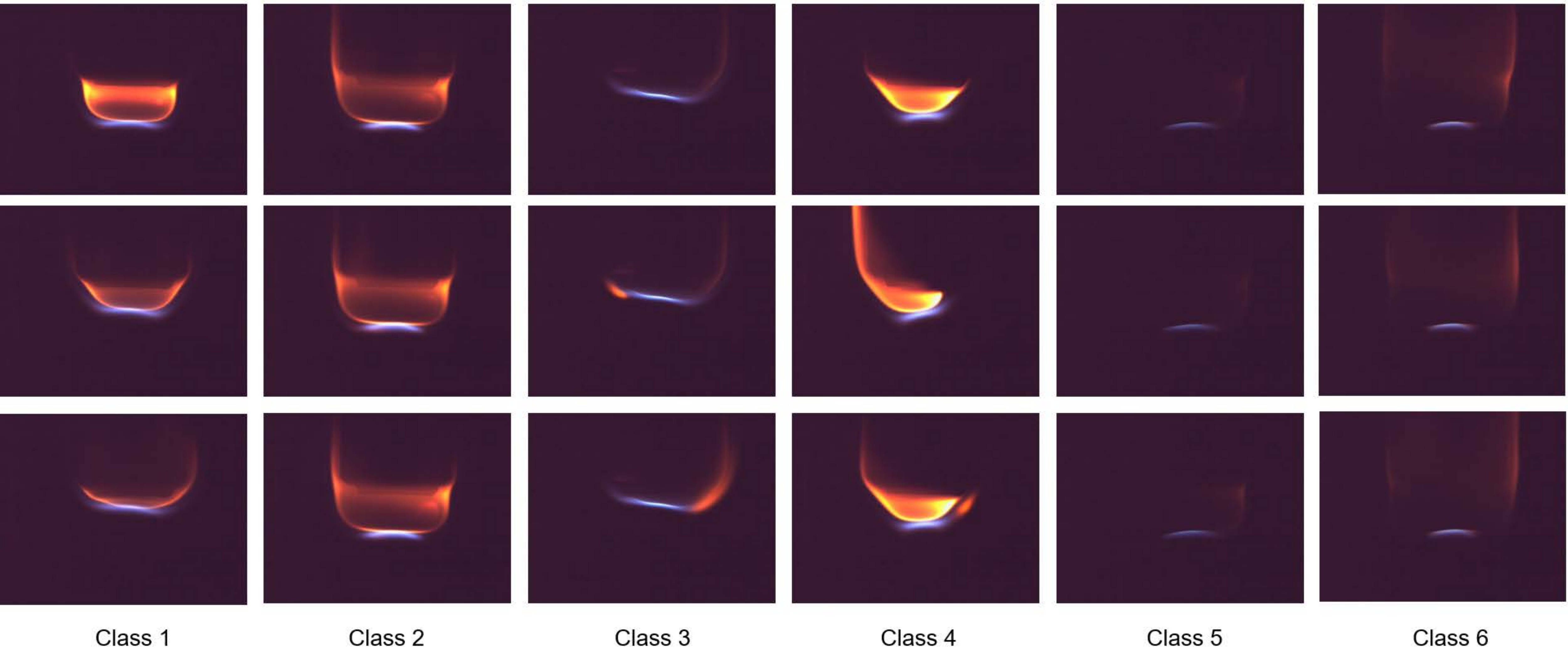}
\caption{Visible images of six flame classes}
\label{fig2}
\end{figure}

\begin{figure}[hbt!]
\centering
\includegraphics[width=1.\textwidth]{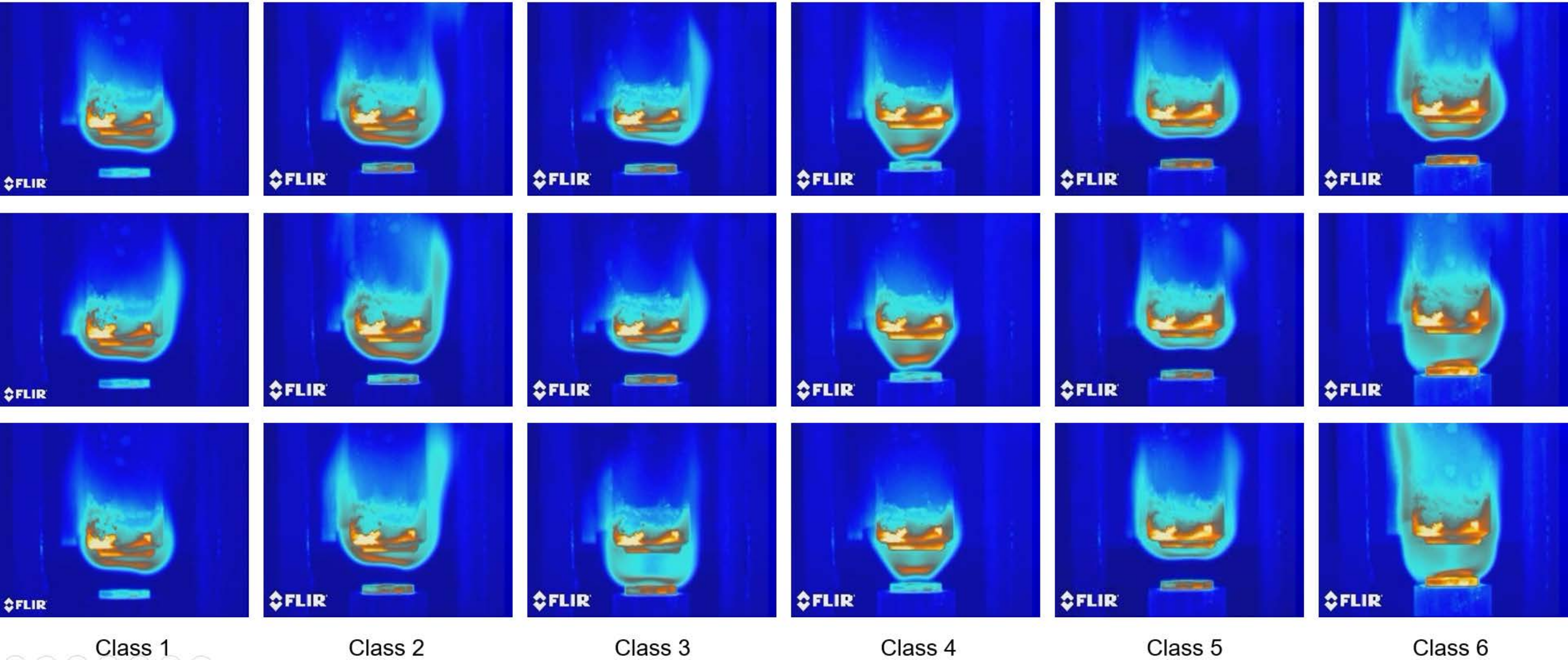}
\caption{IR images of six flame classes}
\label{fig3}
\end{figure}

Sample images from the CMOS (visible-band) and the IR cameras are shown in Figs.~\ref{fig2} and ~\ref{fig3}. Each camera was used for about 20 seconds to take images of every flame class.

After image collection, 20 images per state were randomly picked to construct a training set, and another 20 images were used to build a validation set. To thoroughly evaluate the performance of few-shot learning models, a much larger test set, containing 400 images per state, was built. It is notabed that only the images in the training set, i.e., 20 images per state, were used in order to tune the learnable parameters in models. Compared to the previous reports (Table ~\ref{table1}), it is a much less image amount.

\subsection{Few-shot learning algorithms}
In this work, two few-shot learning algorithms were explored, the first being is Siamese Network (SN) ~\cite{melekhovSiameseNetworkFeatures2016}, which is initially designed to distinguish the similarity of two images. In order to realize the flame state classification, we couple it with k Nearest Neighbors (kNN) algorithm ~\cite{jiangSurveyImprovingKNearestNeighbor2007}. The second algorithm is Prototypical Network (PN) ~\cite{snell2017prototypical}. In following sections, we briefly introduce the their specific design for the flame state classification task herein.

\subsubsection{Siamese Network coupled with k Nearest Neighbors (SN-kNN)}
Siamese network is a contrastive-learning-style few-shot learning algorithm. The function of SN herein is to learn compact encoding of samples which elaborate the class differences between image samples. To do so, we first label pairs of images from the same class with the value of one, and label pairs of images from different classes as zero, as described in Eq. ~\ref{func1},

\begin{equation}
\label{func1}
label =
\begin{cases}
\begin{split}
1\qquad & s_i,s_j\in class_c  \\
0\qquad & s_i\in class_c, s_j\notin class_c
\end{split}
\end{cases}
\end{equation}
where, $s$ represents the image sample, the subscripts $i,j, c$ respectively represent the sample indexes of two image samples, and a state/class index.

\begin{figure}[hbt!]
\centering
\includegraphics[width=0.9\textwidth]{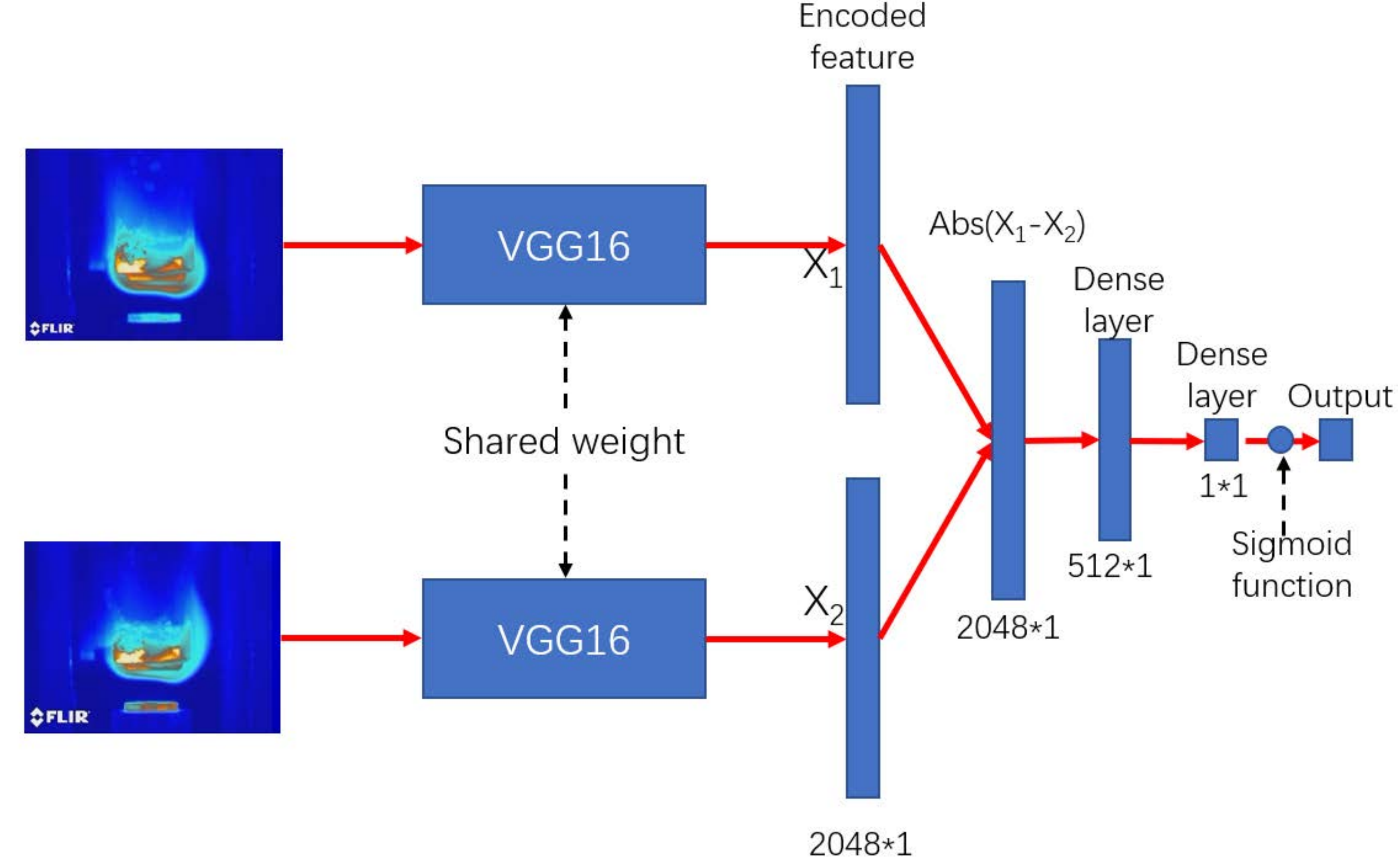}
\caption{Schematic representation of the architecture of the employed Siamese Network}
\label{fig4}
\end{figure}

Then image pairs were fed to SN and the SN output a single value in the range of [0,1] for every pair to mimic the corresponding label of the image pair. The SN architecture through which this was achieved is demonstrated in Fig.\ref{fig4}. It can be mainly divided into two parts, encoding module and prediction module. The encoding module contains two parallel encoding layers, which is used to extract features from images. Herein, we utilized the convolution layers of VGG16 ~\cite{simonyanVeryDeepConvolutional2015} to be encoding layers, of course, any other convolution layers ~\cite{heDeepResidualLearning2016}, even attention layers ~\cite{dosovitskiyImageWorth16x162021} can realize the same function. The two VGG16 used here share the same weights, i.e., they are identical in the terms of architecture and values of learnable parameters. After encoding the two images into two compact features, either of them has a dimension of 2048, the absolute differences of these two features in every entry are fed to the prediction module. The prediction module contains two projection layers (dense layers), which helps map the differences into a single value, and then a sigmoid function was used in order to compress the output into one number in the range of [0,1] , as SN here in fact is tackling a logistic regression task with the binary label of zero or one. The learnable parameters of SN are updated by the gradient-descent back-propagation of binary cross-entropy loss defined by the ground truth and prediction of labels.

\begin{figure}[hbt!]
\centering
\includegraphics[width=0.9\textwidth]{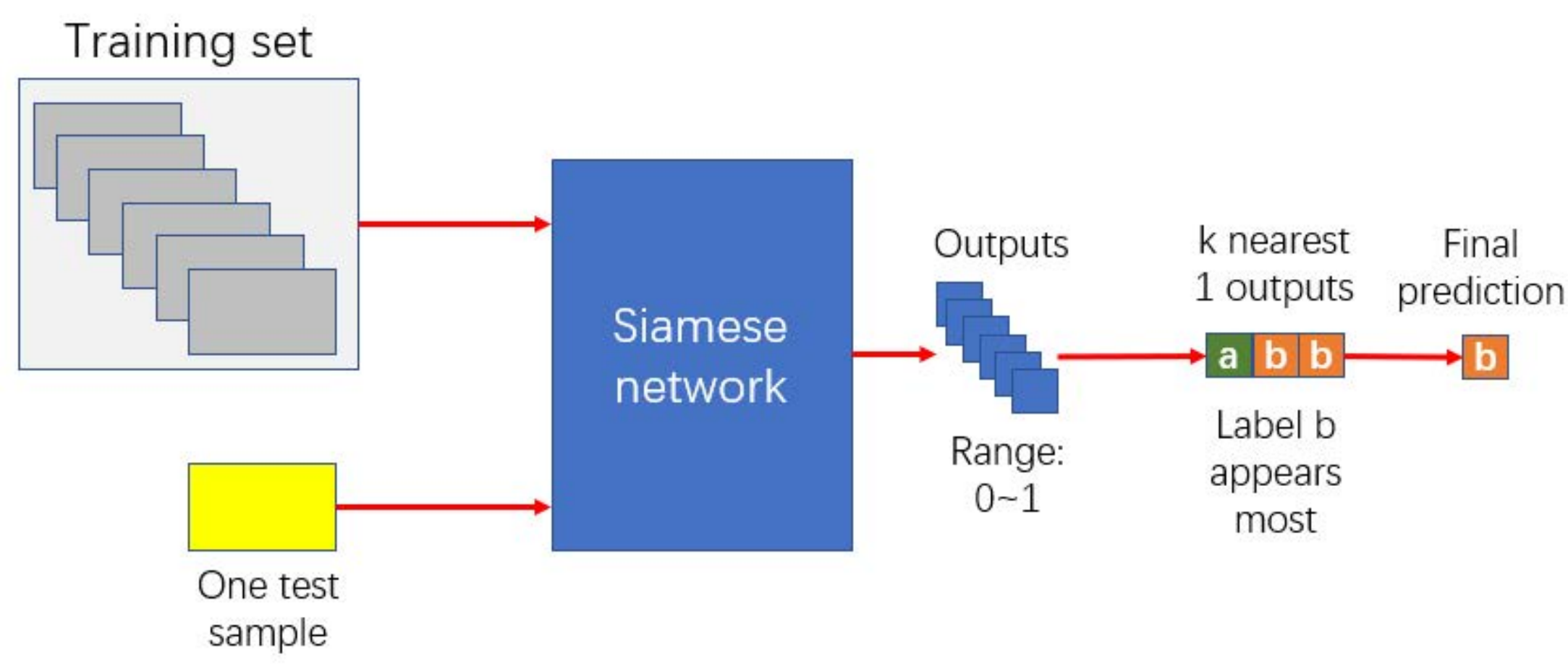}
\caption{The schematic of test flowchart of SN-kNN}
\label{fig5}
\end{figure}

As mentioned above, the SN only indicates whether two images belong to the same classes or not, therefore, we added kNN to complement the classification function. The general idea of kNN is that, for a class-unknown test sample, it should belong to the same class as the plurality of its closest neighbors. In particular, we utilized kNN in the way shown in Fig.~\ref{fig5}. To check the class of one unseen test sample, the test sample is paired with every sample in the training set, and then be fed to SN, the corresponding prediction from the SN indicates the level of similarity between the test sample and a specific training sample. Then, we pick the training samples with the k highest values of predictions to construct a decision set, and then use the class affiliation of most samples in decision set as the class of the test sample. To avoid the ambiguity caused by potentially balanced vote between training samples, the k value used should be an odd number, which is 5 herein.

\subsubsection{Prototypical Network (PN)}
\begin{figure}[hbt!]
\centering
\includegraphics[width=0.9\textwidth]{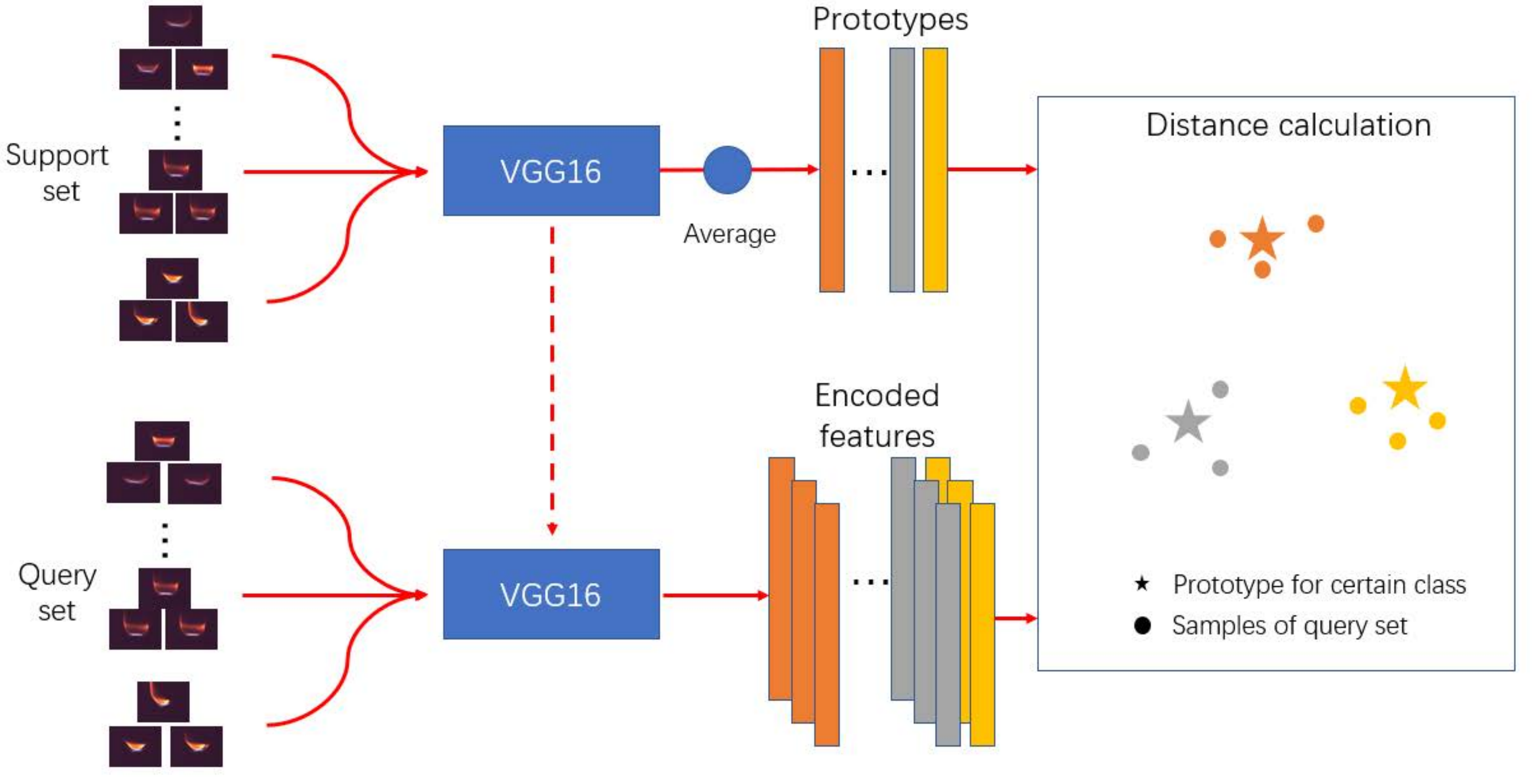}
\caption{The schematic of architecture of Prototypical Network }
\label{fig6}
\end{figure}

In contrast to SN, PN realizes the function of classification by utilizing the concept of prototype. Training of PN can be also divided into two stages, as shown in Fig.~\ref{fig6}. The first stage is constructing prototypes. Randomly picking $n$ samples per class to construct a support set, then encoding these samples of support set into vector-style compact features by an encoding module, which, similarly to the architecture of Fig.~\ref{fig4}, is VGG16, and then averaging the features within the same class to acquire the prototype of the corresponding class (Eq.~\ref{func2}).

\begin{equation}
\label{func2}
    v_{pt,c}=\frac{\sum_{i=1}^{n}v_i}{n} \qquad i\in class_c
\end{equation}

where, $v$ is the encoded features. The subscript $pt$ represents the prototype. As we can see, the prototype is the center of the feature distribution per class, thus, can be regarded as the most comprehensive representative of the given class.

The second stage is updating PN, i.e. generating a query set by picking samples per class from the training set different from the support samples. Again, the features of these query samples are also encoded by VGG16. After that, the Euclidean distances are calculated between the features of the query samples and all prototypes. The distances are transformed to the possibility of the query sample belonging to the specific class by utilizing SoftMax function.

\begin{equation}
\label{func3}
    d_{i,c}=\left\| v_i-v_{pt,c} \right\|_2
\end{equation}

\begin{equation}
\label{func4}
    P_{i,gc}=\frac{exp(-d_{i,gc})}{\sum_{c=1}^{m}exp(-d_{i,c})}
\end{equation}

Where, $d,P$ are respectively distance and probability. The subscripts $gc,m$ respectively demonstrate the given class and total class number. Therefore, the closer the distance between a query sample and a prototype of given class, the higher $P_{i,gc}$ is. Our aim is to make sure that the query sample is closest to the prototype of its own class instead of those of other classes, thus and this is achieved by minimizing the Cross Entropy Loss (CELoss), defined as:
\begin{equation}
\label{func5}
    L=\mathbb{E}_{S_i \sim Q,gc \sim C} \left\{-[y_{i,gc}log(P_{i,gc})+(1-y_{i,gc})log(1-P_{i,gc})]\right\}
\end{equation}

\begin{equation}
\label{func6}
y_{i,gc}=
\begin{cases}
\begin{split}
1\qquad & s_i\in gc  \\
0\qquad & s_i\notin gc
\end{split}
\end{cases}
\end{equation}

Where, $L,Q,C$ are respectively the loss function, the query set, the set of class indexes. As one can see, by minimizing this loss function through backpropagting loss to update the learnable parameters of encoding module, the encoding module will modify the way of embedding images, and tend to make all samples from the same class cluster around the prototype of the class, and expand the distance between samples and prototypes from different classes.

\subsubsection{Training method}
The epoch-batch training method was used to train SN. Bootstrapping sampling ~\cite{mooney1993bootstrapping} was used that, in every batch,  15 images were randomly sampled from the training set. For every image picked, two other images were also picked from the same and a different class respectively, i.e., images generating a positive and a negative contribution for the current sample. Consequently, 15 positive samples plus 15 negative samples make up 30 image pairs with original sampled images. For convenience, we arbitrarily set that one training epoch contains four batches of training.

Episodic training method was used to train PN. The episodic training is also constituted by multiple epochs, and every epoch is made up by multiple episodes. The term episode is the counterpart concept of batch, but training of every episode is executed by a support set and a query set of training set rather than directly the training set. In the training of one episode, we randomly picked five samples per class from the original training set as a support set and did that again from the remaining training samples to construct a query set. Hence, both the training-support set and training-query set had 30 samples. Similar to SN, we also set the number of episodes to four for one epoch training. 

The validation set was also processed in the same way we processed the training set for each algorithm. Once the performance of models on the validation set was improved, the model was automatically saved.

\section{Results and analysis}
\subsection{Image transformation}
In practice, the distance and relative location between the camera and the burner cannot be guaranteed to be identically same during the combustion monitoring process. To avoid models treating the irrelevanted information,, such as the absolute size of flame or burner, flame location in the image, etc. to be meaningful features, we trained the model by utilizing image variants acquired from a series of deliberately designed image transformations. Moreover, image transformation is also a kind of data augmentation that helps avoid model overfitting in the small training set. 

\begin{figure}[hbt!]
\centering
\includegraphics[width=1.0\textwidth]{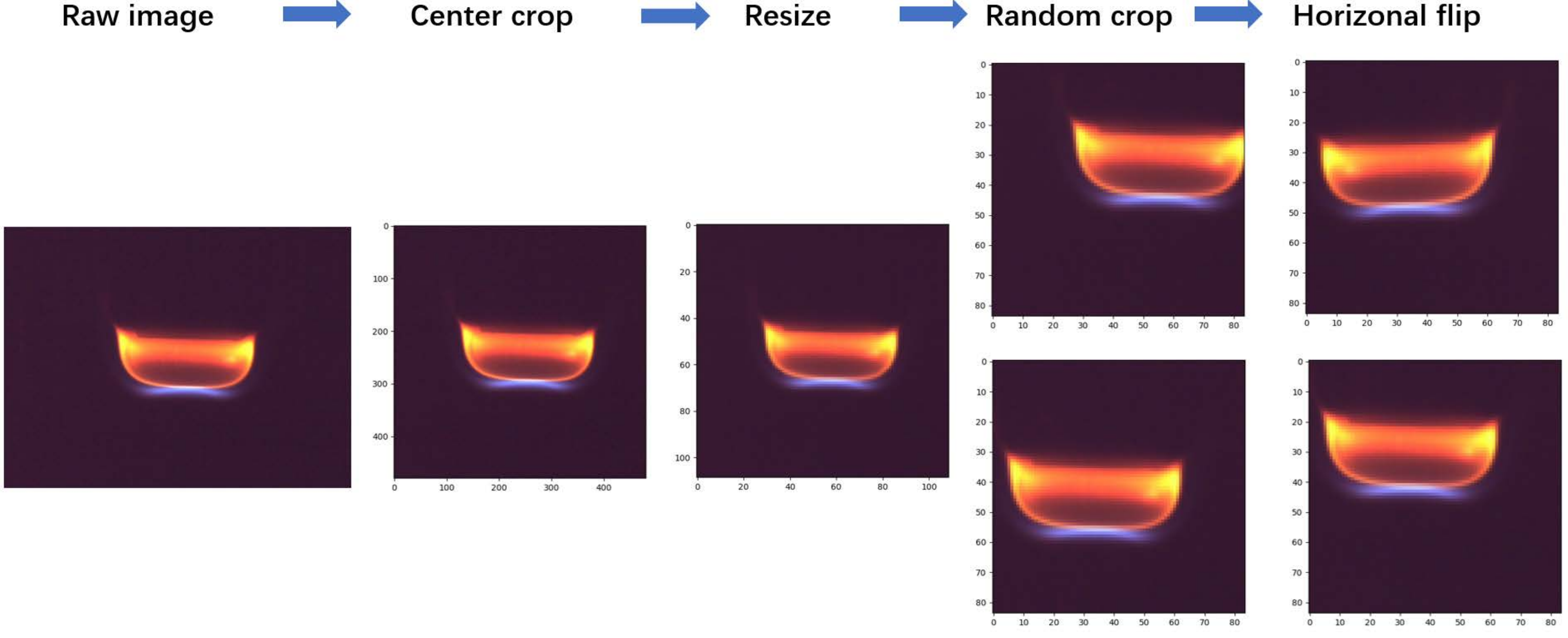}
\caption{The schematic of procedure of Image transformation }
\label{fig7}
\end{figure}

The detailed image transformation procedures are shown in Fig.~\ref{fig7}.
\begin{itemize}
     \item Step 1: Center crop. For the processing convenience of network, the image was cropped to a square with the size of width of original image.
     \item Step 2: Random scale. The image is randomly changed to the size of 1.1-1.5 times of the default input size for the network (84*84 pixels). By doing so, the absolute size of flames from the same class is changeable, and thus will not be regarded as a pattern by the models.
     \item Step 3: Random crop. The image was randomly cropped to the shape of 84*84, therefore, avoiding the position of flame to be learned as a feature. 
     \item Step 4: Horizontal flip. The image was randomly flipped with the possibility of 0.5, because we hypothesize that the flame images taken by the two cameras from opposite sides will be mirror-symmetrical. By doing so, we can augment the dataset.
\end{itemize}

\subsection{Model training}
\label{sec3.2}
\begin{figure}[hbt!]
\centering
\includegraphics[width=.9\textwidth]{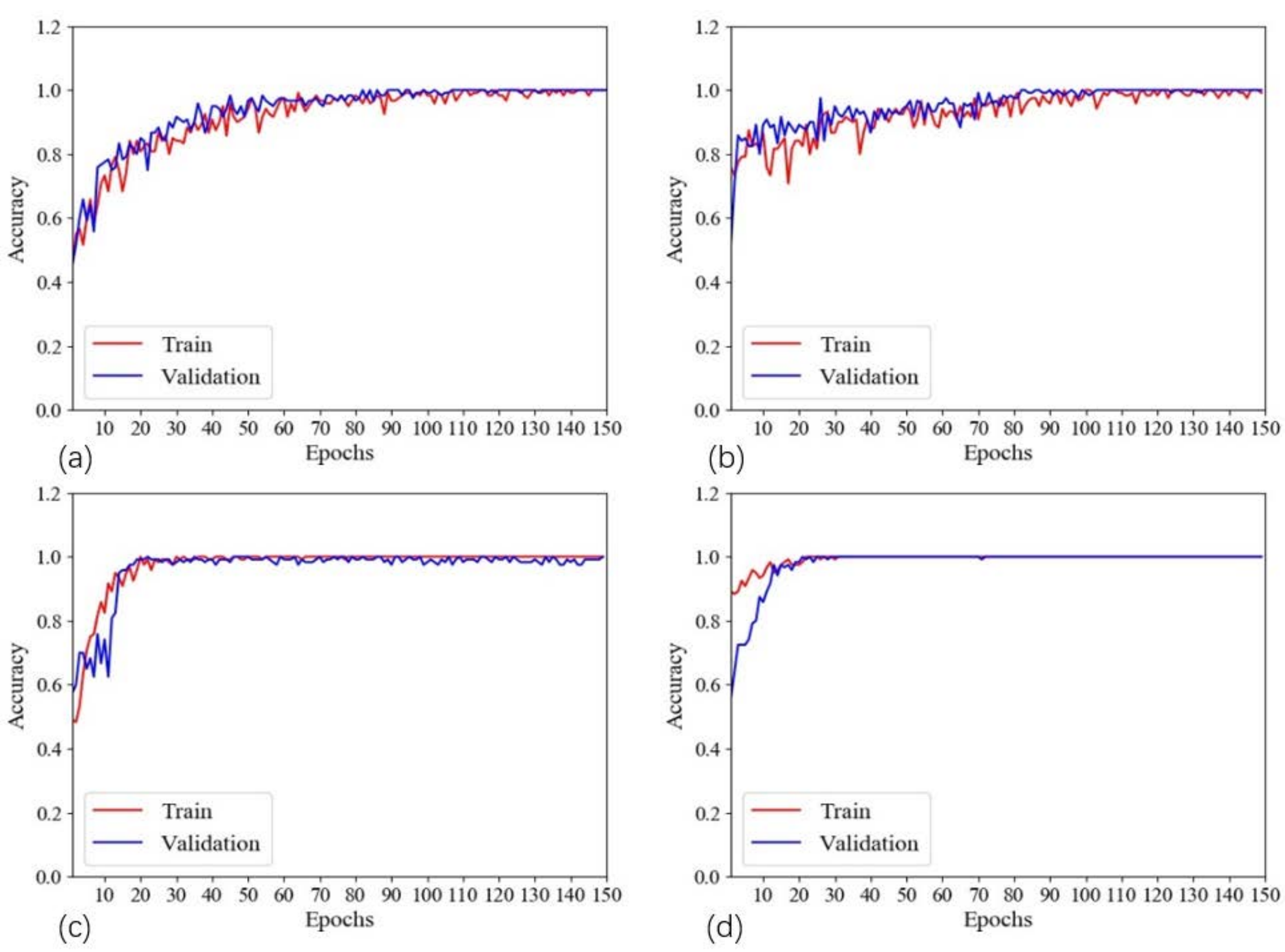}
\caption{Training process of models. (a) SN on IR images (b) SN on visible images (c) PN on IR images (d) PN on visible images}
\label{fig8}
\end{figure}

The training and validation accuracy of two algorithms on two image formats is shown in Fig.~\ref{fig8}. Comparing the convergence time needed for both models, SN needs about 100 epochs to reach the accuracy of 1 (Fig.~\ref{fig8}(a) (b)). whereas PN only needs about 20-30 epochs training (Fig.~\ref{fig8} (c) (d)). The shorter model preparation time of PN is advantageous in industrial applications. This phenomenon is reasonable from the perspective of algorithm architecture. The actual training purpose of SN is to contrast the differences or similarities between samples in each batch, which makes the training twistly, since it may find various perspectives to contrast between individual samples in each batch, but finally, has to compromise to a way that suitable for majority of samples in whole dataset. However, in PN, the purpose of distinguishing differences and similarities between individual samples is transformed to be distinguishing differences between samples and representative prototypes, the latter is a statistical center that more stable than individual sample. Moreover, the combination between prototypes and samples are much fewer than the combinations of individual samples, which further stabilizes the training process.   

Although SN needs similar amount of iterations to converge for both image formats (visible and IR), it is shown in Fig.~\ref{fig8} that the accuracy curve on visible images is flatter (Fig.~\ref{fig8} (b)), which indicates a faster convergence process. As for PN, this trend is even more apparent when comparing Fig.~\ref{fig8} (c) to Fig.~\ref{fig8} (d). Meanwhile, it is easier to get consistent training and validation accuracy on visible images (Fig.~\ref{fig8} (d)). In contrast, training on IR images has more fluctuation after reaching a prediction accuracy of 1 (Fig.~\ref{fig8} (c)). The observations above imply that optimization of the two models is achieved faster by using visible images. The reason can be inferred from comparing raw data from visible and IR images (Figs.~\ref{fig2},~\ref{fig3}), which do show better consistency for the visible samples from the same class and discernible differences between classes.  IR images, on the other hand, have much more (pseudo) color and texture information, and the differences between some classes are not clear enough to make a reasonable classification. For example, some samples of class 1, class 2, class 3, and class 5 in Fig.~\ref{fig3} look similar. To some extent, the IR image has a worse quality than visible image, which makes models learn harder.

The general consistent convergence trends on training and validation set indicates that no overfitting happens, which demonstrates that both network architectures can realize proper training of deep neural network with very limited data.

\subsection{Model testing}
\subsubsection{Prediction performance}
We calculated precision, accuracy, recall, and F1-score in order to assess the performance of models. These metrics are defined as follows:
\begin{equation}
\label{func7}
    A_c=\frac{TP_c+TN_c}{TP_c+TN_c+FP_c+FN_c}
\end{equation}
\begin{equation}
\label{func8}
    P_c=\frac{TP_c}{TP_c+FP_c}
\end{equation}
\begin{equation}
\label{func9}
    R_c=\frac{TP_c}{TP_c+FN_c}
\end{equation}
\begin{equation}
\label{func10}
    F1_c=\frac{2*P_c*R_c}{P_c+R_c}
\end{equation}

where, $A_c, P_c, R_c, F1_c$ indicate the accuracy, precision, recall, and F1-score for particular class c, respectively. For particular class c, we can categorize the sample that belongs to this class as a positive sample; otherwise, the sample is regarded as a negative sample. $TP$ i.e., true positive, means the amount of correctly classified positive samples. $TN$  i.e., true negative, means the amount of correctly classified negative samples. $FP$, i.e., false positive, means the amount of negative samples misclassified as positive . $FN$, i.e., false negative, means the amount of positive samples misclassified as negative. To compare the model test performance comprehensively, we used the macro-average method, i.e., the average of the metric for every class as the final metric.

Table ~\ref{table3} summarizes the metrics of all models. In terms of algorithm performance, PN works worse on IR images, but this performance is still decent, with precision equaling to 0.957, accuracy, recall, and F1-score about 0.953. Besides, PN and SN-kNN have equivalent performances on visible images. When it comes to the impact of image format, IR images, obviously depressed the performance of PN. However, they did not affect adversely the performance of SN-kNN. Actually, for SN-kNN, IR images even demonstrate a slight advantage over visible ones.

\begin{figure}[hbt!]
\centering
\includegraphics[width=.9\textwidth]{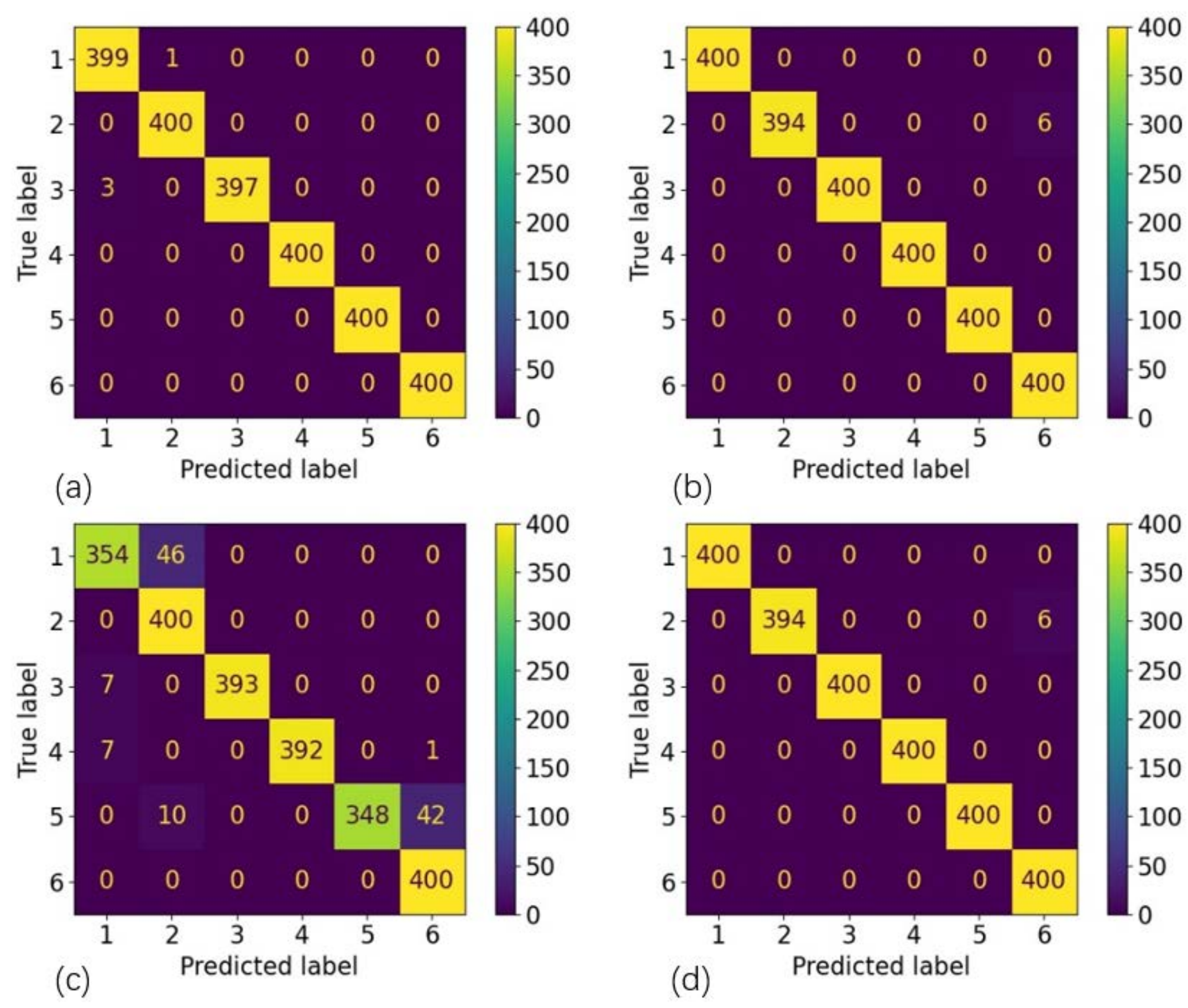}
\caption{Confusion matrices of four models (a) SN on IR images (b) SN on visible images (c) PN on IR images (d) PN on visible images}
\label{fig9}
\end{figure}

\begin{table}
\centering
\caption{Test performance of models}
\label{table3}
\begin{tabular}{|c|c|c|c|c|c|}
\hline
\textbf{Algorithm} & \textbf{Data format} & \textbf{Precision} & \textbf{Accuracy} & \textbf{Recall} & \textbf{F1-score} \\ \hline
\multirow{2}{*}{PN} & IR images & 0.9570 & 0.9529 & 0.9529 & 0.9528 \\ \cline{2-6} 
 & visible images & 0.9975 & 0.9975 & 0.9975 & 0.9975 \\ \hline
\multirow{2}{*}{SN-kNN} & IR images & 0.9983 & 0.9983 & 0.9983 & 0.9983 \\ \cline{2-6} 
 & visible images & 0.9975 & 0.9975 & 0.9975 & 0.9975 \\ \hline
\end{tabular}
\end{table}
Further performance details are revealed by the confusion matrices of Fig.~\ref{fig9}. The confusion matrix provides the information about the classification distribution for a given class. For example, the first row of Fig.~\ref{fig9} (a) tells that except one image of class 1 was misclassified to class 2, SN-kNN correctly classified all the other 399 samples.

\begin{figure}[hbt!]
\centering
\includegraphics[width=.9\textwidth]{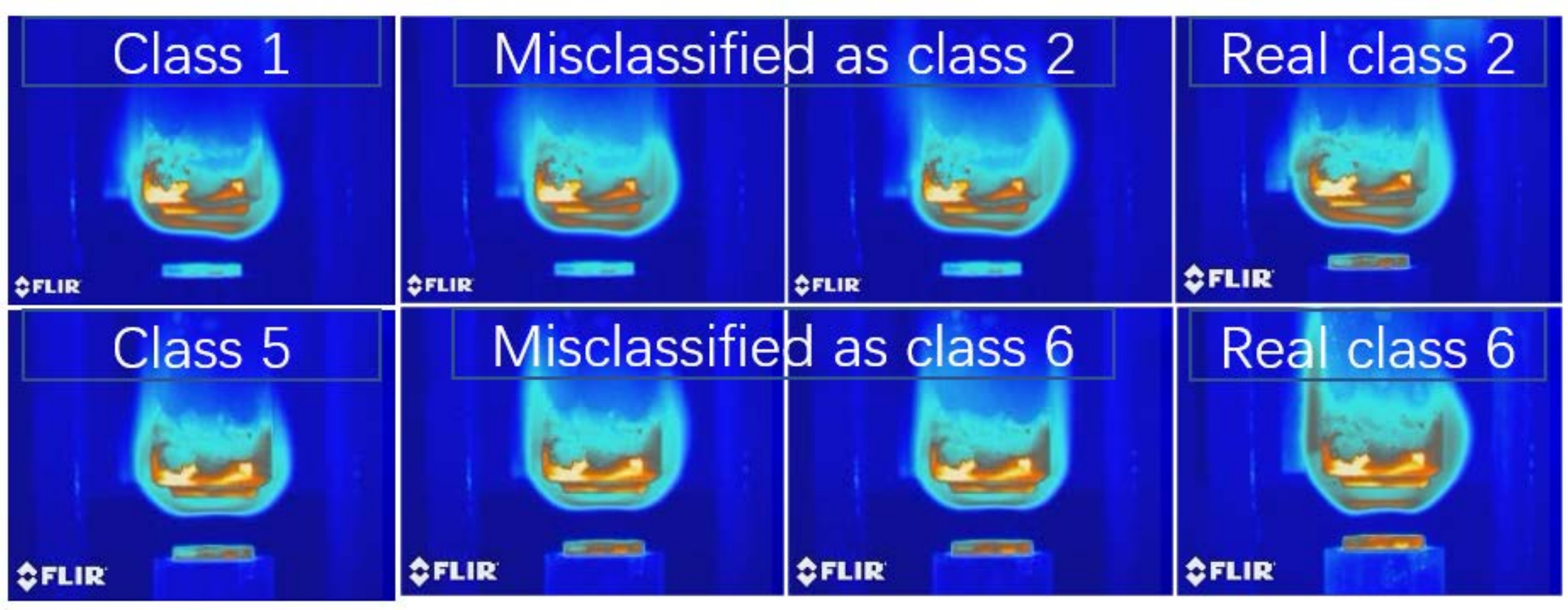}
\caption{The IR images misclassified by PN and the typical images from correct and misclassified classes}
\label{fig10}
\end{figure}

In general, except for the combination of PN and IR images ( Fig.~\ref{fig9} (c)), all the other combinations work well. As for the case of PN working on IR images, about 11.5\% samples of class 1 and 10\% samples of class 5 were misclassified to class 2 and class 6, respectively. Fig.~\ref{fig10} displays some misclassified samples and typical images from the original classes and misclassified classes, it tells that the misclassified images from class 1 and class 5 are visually like the typical images of class 2 and class 6, so it is reasonable that misclassification has happened. As for SN-kNN, it performs almost perfectly on IR images ( Fig.~\ref{fig9} (a)), only very few samples from class 1 and class 3 were misclassified.

\begin{figure}[hbt!]
\centering
\includegraphics[width=.9\textwidth]{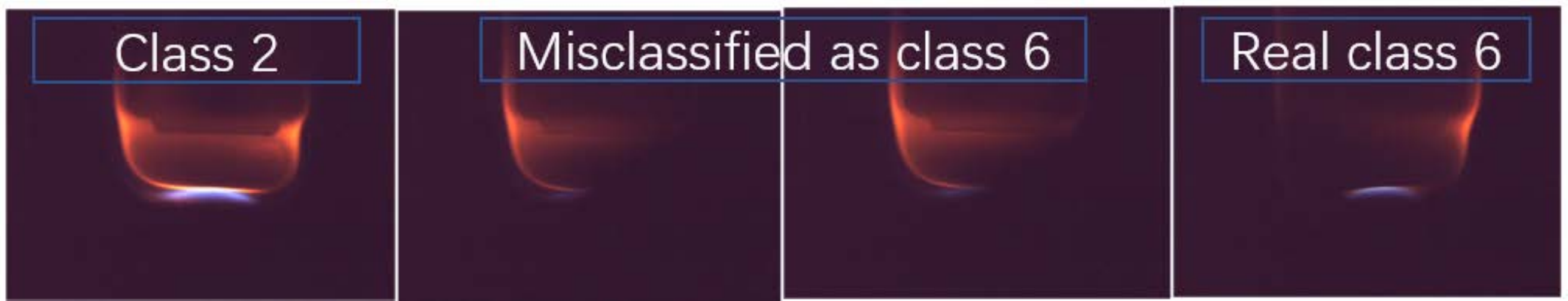}
\caption{The visible images misclassified by PN and the typical images from correct and misclassfied classes }
\label{fig11}
\end{figure}

As for the performance on visible images( Fig.~\ref{fig9}  (b), (d)), both models misclassified the same six images of class 2 to class 6. Fig.~\ref{fig11} displays the two misclassified samples and typical images of class 2 and class 6. As one can see, the typical image of class 2 has an unbroken flame structure. However, these misclassified visible images are broken on the right side of the flame, which makes them more like the typical image of class 6 after horizontal flipping operation. Unfortunately, this horizontal flipping is the transformation utilized in initial data processing, as shown in Fig.~\ref{fig7}. In addition, these six misclassified images were consecutively captured in experiments and coincidentally not sampled by training set. To some extent, these six images can be regarded as outliers of the training set distribution, and consequently cause the confusion of models in the test process. Thereby, it is notable that although few-shot learning algorithms can realize training of deep learning network with little data, models cannot make plausible inference if the relevant information does not conveyed by the training set, as what happens to any other supervised learning algorithm. Apert from these six outliers, both SN-kNN and PN work perfectly on visible images, but both models do demonstrate more misclassifications on IR band. This confirms that IR images seem less suitable for the task considered here, which is consistent to the discussion in Section \ref{sec3.2}. 

The lower quality of IR images harmed the performance of PN, although this did not happen to SN-kNN. This difference in the performance between SN-kNN and PN is attributed to the interaction between the mechanism of data management of the two algorithms and the characteristics of IR images. Specifically, PN needs to summarize the characteristics of samples to construct representatives, i.e., prototypes. However, since IR images have more pseudo-color patterns inside the class, it is challenging for PN to generate the protypes that can represent the characteristics of their own class well . Additionally, the similar appearance of flames from different classes makes this task even harder. In contrast, for SN, there is no need to summarize information in order to generate prototypes. Instead, the algorithm tries to highlight the detailed differences between similar samples from different classes. Meanwhile, kNN does not care about the global distribution of samples for the whole class, instead, only the the local distribution around the test sample is significant. This contrastive learning style and decentralized mechanism makes SN-kNN more robust than PN.

\subsubsection{Model inference speed assessment}
In order to simulate the actual combustion monitoring scenario, we fed the images in the validation set to models in a random sequence frame by frame. The test was executed on an ordinary laptop, whose GPU is GTX1650Ti, CPU is Intel Core i7-9750H. Table ~\ref{table4} shows the total inference time for 120 images, corresponding inference time per frame, and Frame Per Second (FPS).

\begin{table}[]
\centering
\caption{Inferece speed test results}
\label{table4}
\begin{tabular}{|c|c|c|c|c|}
\hline
\textbf{Algorithm} & \textbf{Data format} & \textbf{\begin{tabular}[c]{@{}c@{}}Total inference time \\ for 120   images (ms)\end{tabular}} & \textbf{\begin{tabular}[c]{@{}c@{}}Inference time\\  per frame (ms)\end{tabular}} & \textbf{FPS} \\ \hline
\multirow{2}{*}{PN} & IR images & 5687.5 & 47.40 & 21.10 \\ \cline{2-5} 
 & visible images & 13984.38 & 116.54 & 8.58 \\ \hline
\multirow{2}{*}{SN-kNN} & IR images & 163046.88 & 1358.72 & 0.74 \\ \cline{2-5} 
 & visible images & 276250.00 & 2302.08 & 0.43 \\ \hline
\end{tabular}
\end{table}

Table ~\ref{table4} shows that PN needs 47.40 ms and 116.54 ms to classify one IR image or a visible image respectively, while SN-kNN needs much more time, specifically 28.66 times (IR) or 19.75 times (visible) larger than the corresponding times of PN. This vast difference is mainly caused by algorithm structure. In the test process, the whole training set were used by SN-kNN, which is 120 images in total. However, only six prototypes were needed for PN. Moreover, the prototypes can be recorded in advance, therefore, PN only needs to encode one test sample, but the temporal structure of SN asks that all training samples go through encoding for every test. It is also noted that both few-shot-learning algorithms needed more time to process visible than IR images, because longer reading and transformation times were needed for visible images as they had a larger original size.

\subsection{Feature visualization}
In order to better understand the characteristics of the employed image formats and algorithms, we used t-Distributed Stochastic Neighbor Embedding (t-SNE) algorithm ~\cite{kobak2019art}, a dimensionality reduction algorithm, to map highly-dimensional features encoded by encoders to 2D features. The generated 2D features can then offer a visual, 2-D perception of class clustering. t-SNE is an unsupervised algorithm, which means no label information is needed during feature mapping, it has the advantage of maintaining the relative distance relationship between features after dimensionality reduction, thus allowing to visually observe structure of datasets.

\begin{figure}[hbt!]
\centering
\includegraphics[width=1.0\textwidth]{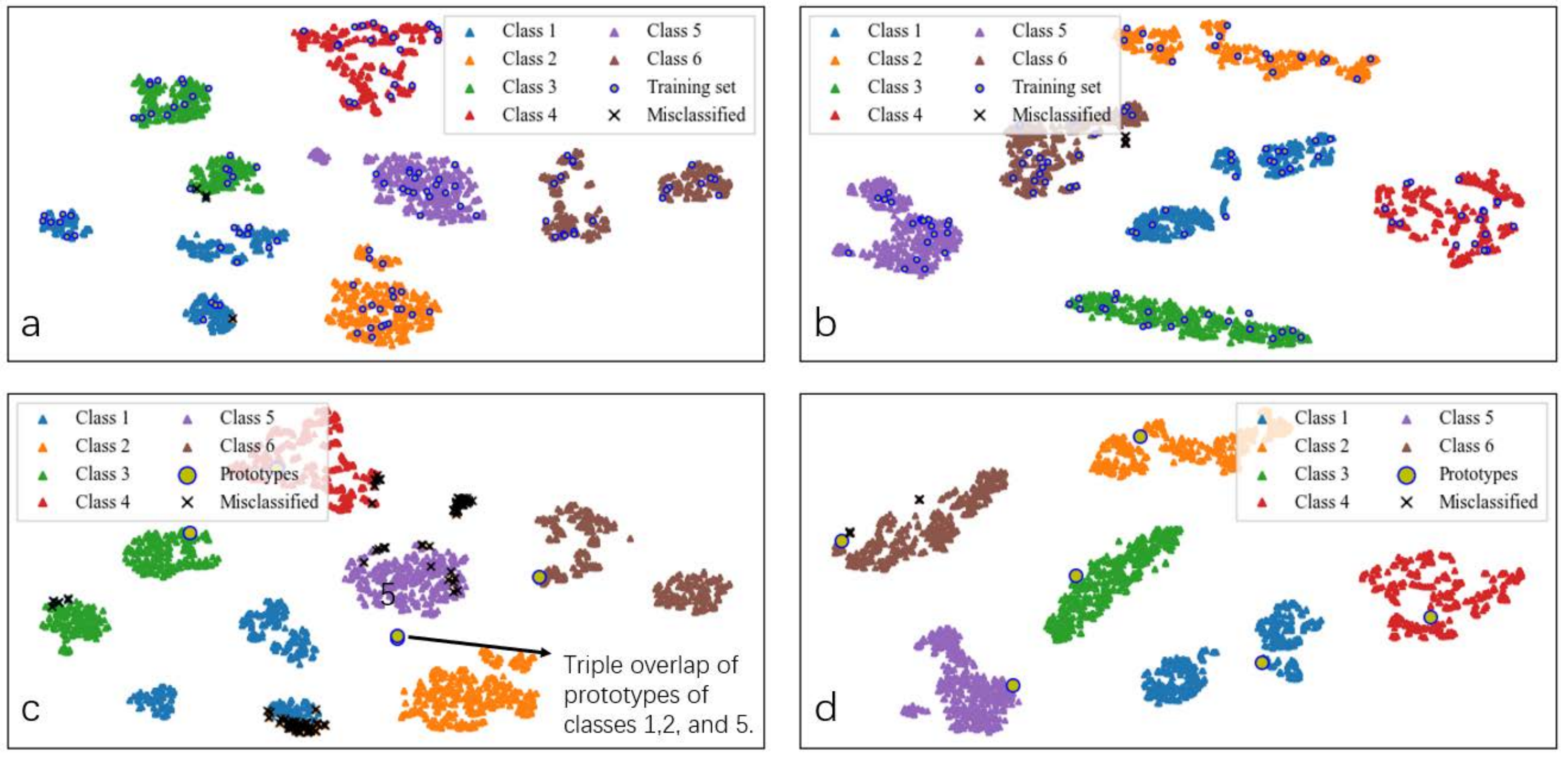}
\caption{Visualization of model-encoded features (a) SN on IR images (b) SN on visible images (c) PN on IR images (d) PN on visible images}
\label{fig12}
\end{figure}

The visualization results of 2D features generated by t-SNE are displayed in Fig.~\ref{fig12}. As shown there, for both visible and IR images, the feature distributions within a specific class are similar for both SN and PN (Fig.~\ref{fig12} (a)(c) for IR images or (b)(d) for visible images). For example, irrespectively for SN or PN, the IR images of class 1 are encoded into three clusters, and one of the clusters can be further divided into two sub-clusters. This phenomenon indicates that both algorithms learn to encode the samples inside the same class in similar ways. However, SN and PN generate different relative positions between classes, e.g, the small cluster belongs to class 5 is encoded in a position close to class 3 by SN (Fig.~\ref{fig12} (a)), whereas, the same small cluster is guided to the position neighbors class 6 by PN (Fig.~\ref{fig12} (c)), which in turn leads to the misclassification of this small cluster.

It can also be observed that IR images have more clusters inside the same class than visible images. E.g. IR images of class 1 (Fig.~\ref{fig12} (a)(c)) develop in three clusters, and class 3, 5, and 6 in two, while for visible images (Fig.~\ref{fig12} (b) (d)), only class 1 can be divided into two or three clusters roughly. The phenomenon of multiple clusters means that the flames generated by the same reactants have multiple significantly distinguishable features/patterns. Also, because of the image quality concerns that we discussed above, the cluster formations of Fig.~\ref{fig12} indicate that IR images have more diverse patterns inside the class, while visible images are more consistent ones. 

Because of the consistent patterns generated by visible images, the PN prototypes generated on visible images are far from each other and inside the cluster of test samples (Fig.~\ref{fig12} (d)), so that PN can work almost perfectly on visible images. However, since the IR images are more elusive, PN cannot generate perfectly separated prototypes, instead, the prototypes of class 1, class 2, and class 5 overlapped, as pointed out in Fig.~\ref{fig12} (c).  This is why 46 images of class 1 (Fig.~\ref{fig9} (c)) are misclassified as class 2. Since prototypes are not contrastive enough, the small cluster of class 5 neighboring a cluster of class 6 is misclassified to class 6, as the prototype of class 6 is closer to this small cluster. Meanwhile, from Fig.~\ref{fig12} (a), which displays the locations of training samples, we find that this small cluster has never been sampled to be part of training set, which in turn caused the generation of biased prototype. However, for SN-kNN, with the support from a broad distribution of samples in the training set, SN works much more robustly on IR images.

\section{Conclusion}
Two few-shot-learning-based algorithms were introduced and developed for combustion state monitoring from visible or IR images, namely Siamese Network coupled with k Nearest Neighbors (SN-kNN) and Prototypical Network (PN). From the results and analysis, we can conclude the following: 

1. Both algorithms are capable of learning to classify flame states from only 20 visible or IR images per class and demonstrate decent performance. More specifically, the worst performance happened for PN on IR images for which the precision was 0.957, and accuracy, recall, and F1-score were about 0.953. Through visualization of features, we found that the the reason for this was that the prototypical network was confused by IR images and could not provide distinguishable prototypes for some classes.  SN on the other hand could utilize the regional distribution of training samples in order to support flame-state classifications, which let to more robust performance. 

2. Visible images were more suitable for flame-state classification than IR images, since both SN-kNN and PN were easier to converge on visible images and the worst test performance happened on IR images. Through direct observation of images and visualization of learned features, we realized that the reason behind this was that visible images had more contrastive differences between flame states and also exhibited consistent patterns inside each class.  This clustering property made visible images easier to classify. 

3, PN only needed about 30 epochs to train, 47.4 ms to classify one IR image, and 116.5 ms to classify one visible image.  In contrast, SN-kNN needed much longer time in training and inference. These advantages in preparation and inference time of PN, may be favoured in efficiency-demanding industrial applications, despite the  superior performance of SNN-kN, especially for IR images.

\section*{Acknowledgement}
DCK would like to acknowledge partial support from Khalifa University of Science and Technology under project RC2-2019-007.
\bibliographystyle{unsrt}  
\bibliography{reference}

\end{document}